\documentclass{article}

\usepackage{arxiv}
\usepackage[numbers,sort&compress]{natbib}

\usepackage[utf8]{inputenc}
\usepackage[T1]{fontenc}

\usepackage{amsmath,amsfonts}
\usepackage{graphicx}
\usepackage{booktabs}
\usepackage{multirow}
\usepackage{adjustbox}
\usepackage{enumitem}
\usepackage{microtype}
\usepackage{xurl}
\usepackage[hidelinks]{hyperref}

\DeclareMathOperator*{\argmin}{arg\,min}

\newcommand{\method}{RPC}
\newcommand{\stgnf}{STG-NF}
\newcommand{\daflow}{DA-Flow}
\newcommand{\auc}{AUROC}
\newcommand{\meanstd}[2]{\ensuremath{#1_{\scriptstyle \pm #2}}}
\newcommand{\bestmeanstd}[2]{\ensuremath{\mathbf{#1}_{\scriptstyle \mathbf{\pm #2}}}}

\title{Reliability-Aware Prototype Calibration for Frozen Pose-Flow Video Anomaly Detection}

\date{}

\author{
\href{http://orcid.org/0000-0003-3045-9798}
{\includegraphics[scale=0.06]{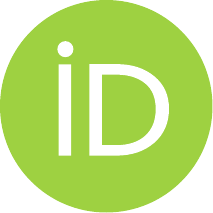}\hspace{1mm}Ning Dong}\thanks{Equal contribution.} \\
School of Information Engineering\\
Suqian University\\
Suqian, China\\
\texttt{dongning@squ.edu.cn}
\And
\href{http://orcid.org/0000-0003-2348-5082}
{\includegraphics[scale=0.06]{orcid.pdf}\hspace{1mm}Yingna Su}\footnotemark[1] \\
School of Information Engineering\\
Suqian University\\
Suqian, China\\
\texttt{suyingna@squ.edu.cn}
\AND
Xin Dong \\
School of Electronic \& Information Engineering\\
Nanjing University of Information\\
Science and Technology\\
Nanjing, China
\And
Ziyun Jiao \\
School of Computer Science and Engineering\\
University of Electronic Science\\
and Technology of China\\
Chengdu, China
\AND
Xinnian Guo \\
School of Information Engineering\\
Suqian University\\
Suqian, China
\And
\href{https://orcid.org/0009-0009-0451-2162}
{\includegraphics[scale=0.06]{orcid.pdf}\hspace{1mm}Zhuangzhuang Pan}\thanks{Corresponding author.} \\
Institute for Advanced Studies\\
Universiti Malaya\\
Kuala Lumpur, Malaysia\\
\texttt{23078403@siswa.um.edu.my}
}

\renewcommand{\shorttitle}{RPC for Frozen Pose-Flow Video Anomaly Detection}

\begin{document}

\maketitle

\begin{abstract}
Pose-flow video anomaly detectors are attractive for one-class surveillance because they provide likelihood-based rankings for tracked skeleton windows. However, a single likelihood score may hide multimodal normal behavior and be sensitive to pose-observation noise. We study a frozen-detector setting in which the pose-flow backbone, cached skeleton tracks, and evaluation pipeline are fixed. Reliability-Aware Prototype Calibration (\method{}) is a post-hoc score calibration method for this setting. It adds a standardized nearest-prototype deviation in the frozen latent space to the standardized flow score, and uses keypoint confidence only to gate this added geometric evidence. Thus, \method{} preserves the original density signal while correcting the ranking with empirical normal-mode structure under pose reliability. Across two frozen pose-flow backbones and four datasets, \method{} improves frame-level \auc{} in all eight backbone--dataset pairs, with gains ranging from $0.34$ to $4.49$ percentage points and averaging $2.03$ points. Ablation and reliability analyses show that prototype deviation is the main corrective signal, while reliability gating is most useful when pose observations are less trustworthy. These results suggest that lightweight post-hoc calibration can strengthen cached pose-flow systems when retraining or reproducing the full pose pipeline is impractical. Code is available \href{https://github.com/iNing10/RPC}{\texttt{github.com/iNing10/RPC}}.
\end{abstract}


\section{Introduction}
\label{sec:introduction}

Video anomaly detection aims to identify events that deviate from normal patterns in surveillance videos~\citep{sultani2018realworld,wu2025anomaly}. It plays an important role in public space monitoring, transportation safety, and networked surveillance systems~\citep{liu2024generalized,liu2025networking}. In practice, abnormal events are rare, diverse, and often unavailable during training, making one-class learning from normal videos a common and realistic setting~\citep{liu2018future,ruff2018deep}. 
For human-centered scenes, skeleton-based representations provide a compact and task-relevant alternative to raw visual appearance. By focusing on tracked body poses, they largely suppress the influence of background clutter, identity, and illumination, while retaining essential cues of human motion and interaction~\citep{flaborea2024contracting}.

Pose-flow detectors provide a natural backbone for one-class skeleton-based anomaly detection~\citep{hirschorn2022stgnf}. They model normal pose sequences with normalizing flows and rank test windows by likelihood-derived scores, e.g., negative log-likelihood. 
However, likelihood primarily reflects density under the learned model rather than the behavioral abnormality itself. Previous studies have shown that deep generative likelihoods can be misaligned with out-of-distribution semantic detection~\citep{nalisnick2019generative}, and that normalizing flows can overemphasize local statistical regularities instead of capturing the decision boundary between normal and abnormal behavior~\citep{kirichenko2020flows}. This limitation becomes particularly pronounced in surveillance videos. Normal behavior is inherently multimodal~\citep{xing2023partition}. Walking, stopping, turning, interacting, and viewpoint-dependent motion may occupy different regions of the learned latent space. A single negative log-likelihood score does not explicitly assess whether a test skeleton lies close to any empirical mode of normal behavior. This motivates a complementary normal-reference signal that checks whether a test skeleton lies close to empirical modes observed in normal training data. Rather than assuming the frozen latent space to be a perfect semantic manifold, we exploit the empirical geometry exposed by the trained detector to calibrate anomaly ranking.

Pose observation quality introduces another source of ambiguity. Skeletons are produced by upstream pose detectors and trackers rather than by pristine motion sensors~\citep{jin2020wholebody}. Occlusion, truncation, small person scale, motion blur, and identity switches can reduce keypoint confidence or corrupt coordinate estimates. However, low keypoint confidence alone does not indicate anomalous behavior. 
% We therefore use confidence to modulate pose-dependent geometric evidence, treating it as a reliability signal rather than direct evidence of anomaly. 
Using confidence directly as anomaly evidence risks penalizing difficult but normal observations, instead of true abnormal behavior. A more conservative strategy is to let confidence modulate the contribution of geometry-based evidence. This separation is consistent with generalized out-of-distribution detection, where density, distance, and other evidence types provide distinct sources of information~\citep{yang2024oodsurvey}. It also aligns with recent skeleton-based studies that highlight the impact of pose quality on anomaly detection~\citep{delic2025seeker}.

Most recent skeleton-based video anomaly detectors address these challenges by improving the detector itself. Explicit distribution modeling, contracted skeletal kinematics, and semantic action cues improve the representation or the training objective~\citep{flaborea2024contracting}. These detector-centric advances are important, but they leave a practical gap when the detector and the cached pose tracks must remain fixed. In surveillance deployments and retrospective benchmark studies, pose extraction and tracking are expensive, sensitive to implementation details, and difficult to reproduce exactly~\citep{andriluka2018posetrack,jin2020wholebody}. As a result, cached skeleton tracks and released detector checkpoints often become the starting point for further adaptation. In this setting, the anomaly score is often the only practical interface for adaptation. A useful score-level calibration method should therefore preserve the frozen pose-flow detector, recover information compressed by the likelihood score, and exploit pose confidence without turning observation unreliability into an anomaly cue.

We propose Reliability-Aware Prototype Calibration (\method{}), a reliability-aware normal-mode calibration method for likelihood-based pose-flow anomaly scoring. \method{} starts from the observation that a frozen pose-flow pipeline already exposes three complementary quantities, namely a likelihood-derived density score, a latent representation of normal motion, and keypoint confidence values. It summarizes normal training latents into normal behavior prototypes and measures each test window by its deviation to the nearest prototype. The resulting prototype deviation score is standardized and used as an additive correction to the standardized likelihood score, while a confidence reliability gate modulates only this added prototype term. In this way, \method{} enriches the original likelihood ranking with normal-mode structure and pose-observation reliability, without neural retraining or changes to the backbone, cached skeleton tracks, or evaluation pipeline. Across two frozen pose-flow backbones and four datasets, \method{} improves frame-level \auc{} over the reproduced likelihood-only baselines in all eight settings.

In summary, the contributions of this paper are:
\begin{itemize}[leftmargin=*]
  \item We formulate frozen pose-flow adaptation as a score-level calibration problem that separates flow density evidence, empirical normal-mode deviation, and pose-observation reliability.
  \item We introduce a post-hoc prototype deviation score in the frozen latent space and combine it with the standardized likelihood-derived score as a conservative ranking correction, rather than a new generative density model.
  \item We design a reliability-gated calibration rule that treats keypoint confidence as observation reliability, not anomaly evidence, and gates only the added prototype deviation.
  \item We show across two frozen pose-flow backbones and four datasets that \method{} improves frame-level anomaly ranking without backbone retraining. Ablations and reliability-oriented analyses further support the proposed prototype and reliability design.
\end{itemize}

\section{Related Work}
\label{sec:related-work}

\subsection{Skeleton-Based and Pose-Flow Video Anomaly Detection}
\label{sec:related-pose}

Video anomaly detection has been studied through reconstruction, prediction, density estimation, weak supervision, retrieval, and multimodal reasoning, as summarized in recent surveys~\citep{su2024vpe,al2026hierarchical}. In human-centered surveillance scenes, skeleton-based methods provide a motion-focused alternative to appearance-based representations by operating on tracked pose trajectories. They reduce appearance-related clutter, such as scene texture, identity, and illumination variation, while preserving body motion and interaction cues. Their effectiveness, however, depends critically on the quality of the upstream pose estimation and tracking pipeline. Recent skeleton-based detectors improve normality modeling through explicit distribution learning, contracted skeletal kinematics, and semantic action-level priors~\citep{tang2026action}.

Pose-flow detectors are particularly relevant to our setting because they provide likelihood-derived anomaly scores in a one-class training regime. \stgnf{} models pose graph sequences with normalizing flows, and \daflow{} further strengthens this family with dual attention~\citep{hirschorn2022stgnf,wu2024daflow}. These methods primarily improve the detector through representation design or training-time modeling. By contrast, our work considers a complementary post-hoc setting where the pose-flow detector, checkpoint, and cached pose tracks are already fixed. This shifts the focus from training a new detector to calibrating the anomaly ranking produced by an existing frozen detector.

\subsection{Likelihood, Prototype, and Post-hoc Normality Scoring}
\label{sec:related-normality}

Likelihood-based scoring is appealing in one-class detection because it provides a scalar ranking with respect to the normal training distribution. However, density evidence does not necessarily align with semantic abnormality. Deep generative models can assign unintuitive likelihoods to out-of-distribution samples, and normalizing flows may rely on local correlations rather than the desired semantic boundary~\citep{nalisnick2019generative,kirichenko2020flows}. Generalized out-of-distribution detection similarly treats density and distance as related but distinct sources of evidence~\citep{yang2024oodsurvey}.

Prototype, memory, and clustering mechanisms provide a complementary way to expose the structure of normal behavior. Video anomaly detectors have used memory banks, partitioned normal references, and clustering objectives to represent reusable normal patterns~\citep{xing2023partition,qiu2024clustering}. In skeleton anomaly detection, GEPC uses clustered pose graph embeddings as part of its detector design~\citep{markovitz2019gepc}. These works motivate prototype-based normal references, but they typically incorporate the reference structure into training. Our use of prototypes is narrower. The frozen flow score remains the primary density signal, and prototypes provide compact reference statistics for empirical normal modes. This differs from replacing the flow score with a Gaussian model, kNN score, OC-SVM, or isolation score in the latent space. \method{} therefore calibrates an existing pose-flow ranking instead of constructing a standalone latent detector.

\subsection{Pose Reliability and Score Calibration}
\label{sec:related-reliability}

Skeleton anomaly scores are affected by the reliability of the upstream pose estimation and tracking pipeline. Occlusion, truncation, small person scale, motion blur, and tracking switches can perturb keypoints before they are observed by the anomaly detector. Recent methods have begun to account for this uncertainty more explicitly. SeeKer incorporates detector confidence into a keypoint-level density estimator, while multimodal evidential learning models reliability under open-world weak supervision~\citep{delic2025seeker}. These studies highlight the importance of observation quality, but they typically modify the estimator, supervision setting, or modality fusion mechanism. In contrast, \method{} keeps the frozen likelihood estimator unchanged and uses confidence only to modulate the added prototype-based geometric evidence.

This reliability view is related to calibration, but its target differs from classical neural calibration. Standard calibration usually concerns the reliability of posterior probabilities rather than the ranking quality of anomaly scores~\citep{guo2017calibration}. Our target is narrower: score-level calibration for anomaly ranking with a frozen one-class detector. \method{} does not claim posterior calibration and does not infer abnormality from low confidence. Instead, confidence controls how strongly prototype deviation contributes to the final ranking. An unreliable skeleton may still receive a high anomaly score from the frozen likelihood estimator, but its distance to normal behavior prototypes may partly reflect pose error rather than behavioral deviation.

\section{Methodology}
\label{sec:method}

\subsection{Overview}
\label{sec:overview}

\method{} is a post-hoc score-level calibration method for frozen pose-flow detectors. It keeps the backbone, pose cache, and evaluation pipeline fixed, and changes only the window-level anomaly ranking score. A pose pipeline converts tracked human trajectories into skeleton windows. Each skeleton window is denoted by
\begin{equation}
  x=\{(P_t,C_t)\}_{t=1}^{T},
\label{eq:skeleton_window}
\end{equation}
where $T$ denotes the window length, $P_t\in\mathbb{R}^{J\times 2}$ contains the coordinates of $J$ keypoints at frame $t$, and $C_t\in[0,1]^J$ contains the corresponding keypoint confidence scores.

The training set $\mathcal{D}_{\mathrm{tr}}=\{x_i^{\mathrm{tr}}\}_{i=1}^{N}$ contains only normal skeleton windows, whereas the test set $\mathcal{D}_{\mathrm{te}}=\{x_j^{\mathrm{te}}\}_{j=1}^{M}$ may contain both normal and abnormal behavior. A window-level detector assigns a scalar anomaly score $s_j^{\mathrm{te}}=s(x_j^{\mathrm{te}})$ to each test window, with larger values indicating stronger abnormality.

\begin{figure}[t]
\centering
\includegraphics[width=\linewidth]{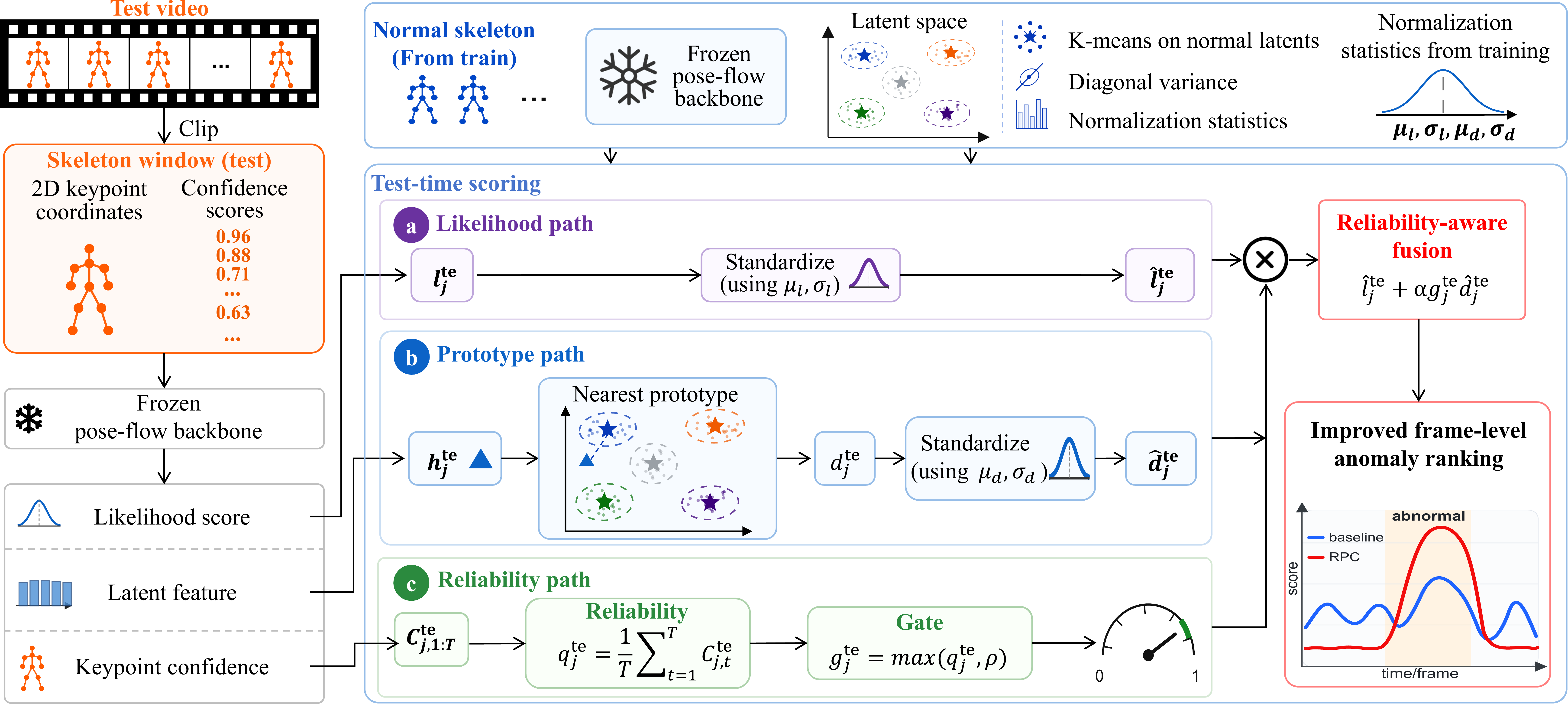}
\caption{Overall framework of \method{}. The frozen backbone remains unchanged, while prototypes, normalization statistics, and reliability gates calibrate the final score.}
\label{fig:framework}
\end{figure}

In our setting, the trained pose-flow detector $f_\theta$ is frozen. Given a skeleton window $x$, the detector provides a latent code $z=f_\theta(x)$ and a likelihood-derived baseline score $s_{\mathrm{flow}}(x)$
\begin{equation}
  s_{\mathrm{flow}}(x)=\ell(x)=-\log p_\theta(x).
\label{eq:flow_score}
\end{equation}
A fixed pooling operator $\phi$ maps $z=f_\theta(x)$ to the feature used for prototype fitting,
\begin{equation}
  h(x)=\phi(z).
\label{eq:latent_feature}
\end{equation}
$h(x)$ is obtained by flattening the latent representation exposed by the frozen detector after its flow-based scoring path, without introducing any trainable projection. Together with the likelihood-derived score and keypoint confidence values, this feature defines the information available for calibration. RPC uses normal training windows to fit prototype statistics and score-normalization constants, and then applies these statistics to calibrate each test-window score. Figure~\ref{fig:framework} summarizes the overall framework.

Since benchmark annotations are frame-level, calibrated window scores are converted into frame-level anomaly scores using the frozen backbone's original aggregation rule. Let $\mathcal{I}(t)=\{j:x_j^{\mathrm{te}}\ \text{is assigned to frame}\ t\}$ denote the test-window indices that contribute to frame $t$. We write this mapping as
\begin{equation}
  a_t=\operatorname{Agg}_{\mathrm{bb}}\left(\{s_j^{\mathrm{te}}:j\in\mathcal{I}(t)\}\right),
\label{eq:frame_aggregation}
\end{equation}
where $\operatorname{Agg}_{\mathrm{bb}}$ denotes the original window-to-frame and multi-person aggregation rule of the reproduced backbone. Performance is measured by frame-level \auc{} computed from $\{a_t\}$ and the frame labels. 

\subsection{Prototype Deviation}
\label{sec:prototype}

% The likelihood score produced by a pose-flow detector measures how well a window fits the learned normal density. Although useful, this scalar score does not explicitly capture the multimodal structure of normal behavior. In surveillance scenes, normal motion may include walking, stopping, turning, waiting, and viewpoint-dependent variations. As a result, a test window may have a moderate likelihood while lying far from the observed normal modes, or it may receive a large negative log-likelihood for reasons that are not semantically abnormal.

To add normal-mode structure to the frozen likelihood ranking, \method{} fits prototypes on normal training features in the latent space exposed by the frozen detector. These prototypes are not interpreted as semantic action classes. They are compact reference statistics used to measure how far a test window lies from the empirical normal modes observed during training.

For each normal training window $x_i^{\mathrm{tr}}$, let $h_i^{\mathrm{tr}}=h(x_i^{\mathrm{tr}})$. We fit $K$ normal behavior prototypes on the normal training features using K-means:
\begin{equation}
  \{c_k\}_{k=1}^{K} =
  \argmin_{\{c_k\}}
  \sum_{i=1}^{N} \min_{1\le k\le K} \|h_i^{\mathrm{tr}}-c_k\|_2^2 .
\label{eq:kmeans}
\end{equation}
Since normal behavior need not occupy one compact region, multiple prototypes provide a more flexible normal reference than a single global center. This step uses only normal training windows and introduces no abnormal labels, neural updates, or additional detector objective. The resulting prototypes are therefore post-hoc statistics of the frozen latent space.

For a test feature $h_j^{\mathrm{te}}=h(x_j^{\mathrm{te}})$, prototype deviation is defined as its distance to the nearest normal behavior prototype. Our main implementation uses a diagonal Mahalanobis distance, which accounts for mode-dependent variance in the latent space and prevents high-variance dimensions from dominating the score merely because of scale. Training features are first assigned to their nearest Euclidean prototype. For prototype $k$ and latent dimension $m$, the diagonal variance is
\begin{equation}
  v_{k,m}=\max\left(\operatorname{Var}\{h_{i,m}^{\mathrm{tr}}: \pi_i^{\mathrm{tr}}=k\}, \epsilon\right),
\label{eq:cluster_variance}
\end{equation}
where $\pi_i^{\mathrm{tr}}$ is the prototype assignment of $h_i^{\mathrm{tr}}$ and $\epsilon=10^{-6}$. For empty clusters, we use the global training variance as a fallback. For a test window $x_j^{\mathrm{te}}$, the raw prototype deviation is then
\begin{equation}
  d_j^{\mathrm{te}}
  =
  d_{\mathrm{proto}}(x_j^{\mathrm{te}})
  =
  \min_k
  \left(
  \sum_m \frac{(h_{j,m}^{\mathrm{te}}-c_{k,m})^2}{v_{k,m}}
  \right)^{1/2}.
\label{eq:prototype_distance}
\end{equation}
We use the diagonal Mahalanobis distance for all reported results since it provides scale-aware nearest-prototype scoring while keeping the calibration post-hoc, gradient-free, and free of additional neural parameters.

Before fusion, the likelihood and prototype terms are standardized using statistics estimated only from normal training windows. The means and standard deviations are computed from $\{\ell(x_i^{\mathrm{tr}})\}_{i=1}^{N}$ and $\{d_{\mathrm{proto}}(x_i^{\mathrm{tr}})\}_{i=1}^{N}$. For each test window $x_j^{\mathrm{te}}$, we compute
\begin{equation}
  \hat{\ell}_j^{\mathrm{te}}=\frac{\ell(x_j^{\mathrm{te}})-\mu_\ell}{\sigma_\ell},
  \qquad
  \hat{d}_j^{\mathrm{te}}=\frac{d_j^{\mathrm{te}}-\mu_d}{\sigma_d}.
\label{eq:standardization}
\end{equation}
% Standardization places density and distance evidence on a common training normal scale. The fusion weight $\alpha$ therefore controls the strength of the prototype correction, rather than compensating for arbitrary numerical ranges. This step is score normalization for ranking, not probability calibration.
This normalization places density and distance on the same normal-training scale, so the fusion weight $\alpha$ controls the strength of the prototype correction rather than compensating for arbitrary score ranges.

% Prototype deviation is an additive correction, not a replacement for the likelihood score. The flow score preserves density evidence learned by the frozen normalizing flow. The prototype term measures distance from empirical normal modes in the same latent space. The final score therefore combines learned density evidence with post-hoc normal mode geometry.

\subsection{Reliability-Gated Score Calibration}
\label{sec:reliability}

% Prototype deviation is computed from pose geometry at test time. With reliable pose observations, a large distance from all normal behavior prototypes can provide useful anomaly evidence. Under unreliable observations, however, the same distance may instead arise from occlusion, truncation, motion blur, small person scale, or tracking error. We therefore use reliability to control the contribution of prototype evidence for each test window, rather than to create an additional anomaly score.
Since the prototype distance is pose-dependent, its contribution should depend on the reliability of the observed keypoints. For a test window
\(x_j^{\mathrm{te}}=\{(P_{j,t}^{\mathrm{te}},C_{j,t}^{\mathrm{te}})\}_{t=1}^{T}\),
let \((C_{j,t}^{\mathrm{te}})_r\) be the confidence of keypoint \(r\) at frame \(t\).
Missing keypoints receive confidence zero when the upstream pose track stores them as missing.
The raw test-window reliability is defined as the clipped mean confidence:
\begin{equation}
  q_j^{\mathrm{te}}=\frac{1}{TJ}
  \sum_{t=1}^{T}\sum_{r=1}^{J}
  \operatorname{clip}\left((C_{j,t}^{\mathrm{te}})_r,0,1\right).
\label{eq:reliability}
\end{equation}
For all retained skeleton windows, at least one keypoint has positive confidence in the window,
so \(q_j^{\mathrm{te}}\in(0,1]\). The candidate \(\gamma=0\) corresponds to ungated prototype fusion,
for which \((q_j^{\mathrm{te}})^0=1\) for all valid windows. 

This quantity uses only confidence values already produced by the pose pipeline and requires no separate uncertainty estimator. The reliability gate for the same test window is
\begin{equation}
  g_j^{\mathrm{te}}=\max\left((q_j^{\mathrm{te}})^\gamma,\rho\right),
\label{eq:gate}
\end{equation}
where $\gamma$ controls the gate curvature and $\rho$ is a reliability floor. Values $\gamma>1$ suppress low confidence windows more strongly, whereas $\gamma<1$ softens the attenuation and keeps the prototype term closer to its ungated form.

The final gated RPC score for test window $x_j^{\mathrm{te}}$ is
\begin{equation}
  s_j^{\mathrm{te}}
  =
  s_{\mathrm{RPC}}(x_j^{\mathrm{te}})
  =
  \hat{\ell}_j^{\mathrm{te}}+\alpha g_j^{\mathrm{te}}\hat{d}_j^{\mathrm{te}}.
\label{eq:rpc_score}
\end{equation}
This calibration rule leaves the standardized negative log-likelihood ungated. As a result, an unreliable test skeleton can still be ranked as anomalous by the frozen density model, but it cannot dominate the calibrated score solely through a prototype distance that may reflect pose estimation error. The gate acts as a conservative controller of geometric evidence, rather than as a confidence-based anomaly detector.

\section{Experiments}
\label{sec:experiments}

\subsection{Datasets and Metrics}
\label{sec:datasets}

We evaluate \method{} on four datasets, which are ShanghaiTech Campus~\citep{luo2017shanghaitech}, UBnormal~\citep{acsintoae2022ubnormal}, and their human-related variants ShanghaiTech-HR and UBnormal-HR~\citep{flaborea2024contracting}. These datasets support skeleton-based video anomaly detection under the one-class setting. For compactness, tables and figures use SHTech for ShanghaiTech and UBN for UBnormal. HR denotes the human-related dataset variant. Table~\ref{tab:dataset-protocols} reports the public statistics for these datasets. These counts are not derived from our cached skeleton tracks.

SHTech contains real campus surveillance videos with multiple scenes, crowded activities, illumination changes, and diverse abnormal events. UBN is a synthetic benchmark with an official validation split, diverse virtual scenes, and anomaly types that differ across splits. These datasets cover complementary regimes. SHTech reflects real surveillance variation, while UBN provides diverse anomaly types and an official validation split.

SHTech-HR and UBN-HR are human-related datasets derived from the corresponding original datasets. They remove nonhuman or appearance-dominated anomalies that skeleton motion cannot reliably explain, such as events dominated by vehicles or fire. This makes the evaluation target better aligned with skeleton-only pose-flow detectors.

\begin{table}[t]
\centering
\caption{Statistics of the evaluated datasets. All numeric values are frame counts. The HR variants follow \citet{flaborea2024contracting} and remove nonhuman or appearance dominated anomalies from the corresponding original datasets.}
\label{tab:dataset-protocols}
\footnotesize
\setlength{\tabcolsep}{3pt}
\begin{adjustbox}{max width=\linewidth}
\begin{tabular}{lccccccl}
\toprule
Dataset & Total & Train & Val & Test & Normal & Abnormal & Public characterization \\
\midrule
SHTech~\citep{luo2017shanghaitech} & 317,398 & 274,515 & -- & 42,883 & 300,308 & 17,090 & 13 real scenes, 130 abnormal events \\
SHTech-HR~\citep{flaborea2024contracting} & 313,212 & 274,515 & -- & 38,697 & 297,090 & 16,122 & HR dataset, 6 nonhuman test videos removed \\
UBN~\citep{acsintoae2022ubnormal} & 236,902 & 116,087 & 28,175 & 92,640 & 147,887 & 89,015 & 29 scenes, 22 anomaly types, official validation split \\
UBN-HR~\citep{flaborea2024contracting} & 234,751 & 116,087 & 28,175 & 90,489 & 147,887 & 86,864 & HR dataset, validation split kept unchanged \\
\bottomrule
\end{tabular}
\end{adjustbox}
\end{table}

All detectors are trained in the one-class setting. Only normal skeleton windows are used to compute frozen detector statistics, fit prototype centers and diagonal variances, and estimate normalization constants. Window-level scores are converted into frame-level anomaly scores by the corresponding frozen backbone evaluation pipeline. We report frame-level area under the receiver operating characteristic curve, denoted as \auc{}.

\subsection{Implementation Details}
\label{sec:implementation}
We use \stgnf{}~\citep{hirschorn2022stgnf} and \daflow{}~\citep{wu2024daflow} as the frozen pose-flow baselines and apply \method{} on top of their reproduced checkpoints. The checkpoint, cached skeleton input, and evaluation pipeline are then kept fixed. RPC uses the final latent variable produced by the frozen flow after the forward transformation. All calibration runs use flattened latent features, diagonal Mahalanobis distance, reliability floor $\rho=0$, and 30 K-means iterations. 

Hyperparameters are selected once for each original dataset and reused for its HR variant without accessing any test data. For UBN, we use the official validation split to select $K$, $\alpha$, and $\gamma$, and then freeze the selected values for both UBN and UBN-HR. Since SHTech has no official validation split in the one-class setting, we construct a training-only proxy validation set from normal SHTech training trajectories, following the use of self-supervised or proxy tasks for VAD model selection~\citep{georgescu2021selfsupervised,yang2025proxy,fung2024modelselection}. Specifically, 20\% of normal training skeleton windows are held out as the proxy set, while the remaining 80\% are used to fit prototypes and estimate normalization statistics.

The SHTech proxy set contains clean held-out normal samples, behavior-preserving corrupted variants with confidence corruption or mild dropout and jitter, and pseudo anomalies generated from the same held-out trajectories. The pseudo-anomaly transformations include velocity drift, acceleration bumps, local limb shifts, temporal warping, speed-up, pose freezing, pose-scale jumps, and track-fragment replacement. 

We select the hyperparameters with the highest validation \auc{} on the SHTech proxy validation set and the UBN official validation split. The finite coordinate search uses the following predefined candidate space:
\begin{align}
K &\in \{2,4,6,8,10,16,32,48,64,80\}, \nonumber\\
\alpha &\in \{0,0.25,0.5,0.75,1.0,1.25,1.5,2.0,2.5,3.0,3.5,4.0,6.0,8.0,10.0\}, \nonumber\\
\gamma &\in \{0,0.25,0.5,0.75,1.0,1.25,1.5,2.0,2.5,3.0,3.5,4.0,6.0,8.0,10.0,12.0,16.0\}. \nonumber
\end{align}
The distance, reliability mode, normalization, pooling rule, and reliability floor are fixed before final evaluation. Table~\ref{tab:configs} records the frozen per-dataset configurations used in the experiments.

\begin{table}[htbp]
\centering
\caption{Hyperparameter configurations of \method{} for each dataset and frozen backbone. Hyperparameters are selected once on SHTech proxy validation or UBN official validation and then reused for the corresponding HR dataset.}
\label{tab:configs}
\footnotesize
\setlength{\tabcolsep}{8pt}
\begin{tabular}{l@{\hspace{1.7em}}ccc@{\hspace{2.8em}}ccc}
\toprule
\multirow{2}{*}{Dataset} & \multicolumn{3}{c}{\stgnf{}} & \multicolumn{3}{c}{\daflow{}} \\
\cmidrule(lr){2-4}\cmidrule(lr){5-7}
 & $K$ & $\alpha$ & $\gamma$ & $K$ & $\alpha$ & $\gamma$ \\
\midrule
SHTech    & 8  & 0.25 & 1.0  & 48 & 0.25  & 1.0  \\
SHTech-HR & 8  & 0.25 & 1.0  & 48 & 0.25  & 1.0  \\
UBN        & 80 & 1.25 & 3.5  & 64 & 4.00  & 3.5  \\
UBN-HR     & 80 & 1.25 & 3.5  & 64 & 4.00  & 3.5  \\
\bottomrule
\end{tabular}
\end{table}

\subsection{Main Results}
\label{sec:main-results}

We first test whether post-hoc score calibration improves frozen pose-flow detectors. Table~\ref{tab:main} compares the reproduced likelihood-only baseline, traditional post-hoc methods applied to the same frozen features, and \method{}. All results are reported as mean $\pm$ standard deviation over five seeds. Baseline denotes the reproduced likelihood-only detector for each frozen backbone. The traditional post-hoc methods include Gaussian modeling, kNN, OC-SVM, and Isolation Forest. Their hyperparameters are selected using the same validation rule as \method{}.

\method{} improves over the reproduced baseline for both frozen backbones and all four datasets. The gains range from $+0.34$ points on \daflow{} SHTech to $+4.49$ points on \daflow{} UBN-HR, with an average gain of $+2.03$ points. \method{} also achieves the best result in every row of Table~\ref{tab:main}. These results indicate that the frozen likelihood score and prototype deviation provide complementary evidence. In this frozen setting, preserving the original likelihood ranking and using prototype deviation as a corrective signal is more effective than replacing the score with a standalone latent-space detector.

\begin{table}[t]
\centering
\caption{Main frame-level \auc{} (\%) results. Values are mean $\pm$ standard deviation over five seeds. Bold values mark the best result in each row. Gain is measured relative to the corresponding baseline.}
\label{tab:main}
\footnotesize
\setlength{\tabcolsep}{3pt}
\begin{tabular}{llccccccc}
\toprule
\multirow{2}{*}{Backbone} & \multirow{2}{*}{Dataset} & \multirow{2}{*}{Baseline} & \multicolumn{4}{c}{Traditional post-hoc} & \multicolumn{2}{c}{Ours} \\
\cmidrule(lr){4-7}\cmidrule(lr){8-9}
 & & & Gaussian & kNN & OC-SVM & IsoForest & \method{} & Gain \\
\midrule
\stgnf{} & SHTech & \meanstd{85.52}{0.32} & \meanstd{84.65}{0.24} & \meanstd{83.22}{0.21} & \meanstd{83.51}{0.32} & \meanstd{84.97}{0.38} & \bestmeanstd{85.89}{0.33} & +0.37 \\
\stgnf{} & SHTech-HR & \meanstd{87.06}{0.28} & \meanstd{86.31}{0.20} & \meanstd{85.00}{0.34} & \meanstd{84.75}{0.32} & \meanstd{86.71}{0.09} & \bestmeanstd{87.56}{0.26} & +0.50 \\
\stgnf{} & UBN & \meanstd{71.81}{0.57} & \meanstd{71.07}{0.10} & \meanstd{73.38}{1.42} & \meanstd{69.99}{0.65} & \meanstd{71.04}{0.54} & \bestmeanstd{74.66}{1.53} & +2.85 \\
\stgnf{} & UBN-HR & \meanstd{72.30}{0.61} & \meanstd{71.47}{0.10} & \meanstd{73.79}{1.24} & \meanstd{70.34}{0.12} & \meanstd{71.26}{0.51} & \bestmeanstd{75.39}{1.57} & +3.09 \\
\daflow{} & SHTech & \meanstd{84.81}{0.47} & \meanstd{84.27}{0.14} & \meanstd{82.87}{0.17} & \meanstd{83.32}{0.30} & \meanstd{84.67}{0.21} & \bestmeanstd{85.15}{0.36} & +0.34 \\
\daflow{} & SHTech-HR & \meanstd{86.46}{0.41} & \meanstd{85.90}{0.17} & \meanstd{84.57}{0.21} & \meanstd{84.48}{0.44} & \meanstd{86.31}{0.10} & \bestmeanstd{87.00}{0.30} & +0.54 \\
\daflow{} & UBN & \meanstd{74.46}{0.69} & \meanstd{72.89}{0.69} & \meanstd{74.49}{0.19} & \meanstd{70.04}{1.15} & \meanstd{71.61}{0.45} & \bestmeanstd{78.54}{0.47} & +4.08 \\
\daflow{} & UBN-HR & \meanstd{75.06}{0.72} & \meanstd{73.37}{0.73} & \meanstd{75.18}{0.29} & \meanstd{71.37}{0.71} & \meanstd{71.99}{0.48} & \bestmeanstd{79.55}{0.39} & +4.49 \\
\bottomrule
\end{tabular}
\end{table}

\subsection{Comparison with Recent Methods}
\label{sec:sota-comparison}

Table~\ref{tab:sota} places the calibrated frozen pose-flow detectors in the context of recent VAD methods. The comparison includes RGB, multimodal, memory, diffusion, language-based, skeleton, pose, and trajectory methods. \stgnf{}+\method{} obtains 85.89 on SHTech and 87.56 on SHTech-HR, while \daflow{}+\method{} obtains 78.54 on UBN and 79.55 on UBN-HR. These results position the calibrated pose-flow detectors favorably among recent skeleton- and pose-based methods.

The table also shows that the calibrated pose-flow detectors remain competitive in a broader comparison that includes methods using richer visual or multimodal cues. This is notable because \method{} does not change the input modality, retrain the detector, or introduce a new representation learner. Instead, the improvements are obtained by calibrating the scores of existing frozen pose-flow backbones. The comparison therefore highlights the practical value of post-hoc calibration for upgrading skeleton-based VAD systems under fixed-backbone and cached-trajectory constraints.
\begin{table}[t]
\centering
\caption{Comparison with recent video anomaly detection methods using frame-level \auc{} (\%). Rows are grouped by method type for readability. Bold and underline mark the best and second best entries in each dataset column. Our rows report the five-seed mean.}
\label{tab:sota}
\footnotesize
\setlength{\tabcolsep}{3pt}
\begin{tabular}{cllcccc}
\toprule
Year & Type & Method & SHTech & SHTech-HR & UBN & UBN-HR \\
\midrule
\multicolumn{7}{l}{\textit{RGB, video, memory, and multimodal methods}} \\
2024 & Video SSL & AED-MAE~\citep{ristea2024sdmae} & 79.10 & -- & 58.50 & -- \\
2024 & Video SSL & MGST~\citep{zhang2024mgst} & 85.10 & -- & -- & -- \\
2024 & Memory and patch & VideoPatchCore~\citep{ahn2024videopatchcore} & 85.10 & -- & -- & -- \\
2024 & Diffusion & DiffVAD~\citep{zhang2024diffvad} & 81.90 & -- & -- & -- \\
2024 & LLM reasoning & AnomalyRuler~\citep{yang2024anomalyruler} & 85.20 & -- & -- & -- \\
2025 & Video SSL & STP~\citep{yang2025proxy} & 81.10 & -- & -- & -- \\
2025 & Video graph SSL & STG-SL~\citep{xing2025stgsl} & 76.80 & -- & -- & -- \\
2025 & Patch diffusion & MA-PDM~\citep{zhou2025mapdm} & 79.20 & -- & 63.40 & -- \\
2026 & LVLM, zero-shot & AnyAnomaly~\citep{ahn2026anyanomaly} & 79.70 & -- & 74.50 & -- \\
2026 & RGB, flow, and pose & SC-MAFC~\citep{wang2026scmafc} & 84.60 & -- & -- & -- \\
\midrule
\multicolumn{7}{l}{\textit{Skeleton, pose, and trajectory methods}} \\
2024 & Skeleton and trajectory & TrajREC~\citep{stergiou2024trajrec} & -- & 77.90 & 68.00 & 68.20 \\
2024 & Skeleton and trajectory & TSGAD~\citep{noghre2024tsgad} & 80.60 & 81.77 & -- & -- \\
2024 & Skeleton transformer & PoseWatch~\citep{noghre2024posewatch} & \underline{85.75} & \underline{87.23} & -- & -- \\
2025 & Skeleton keypoint density & SeeKer~\citep{delic2025seeker} & 85.50 & 86.90 & \underline{77.90} & -- \\
\midrule
\multicolumn{7}{l}{\textit{Ours: score calibration on frozen pose-flow backbones}} \\
2026 & Ours, score calibration & \stgnf{}+\method{} & \textbf{85.89} & \textbf{87.56} & 74.66 & \underline{75.39} \\
2026 & Ours, score calibration & \daflow{}+\method{} & 85.15 & 87.00 & \textbf{78.54} & \textbf{79.55} \\
\bottomrule
\end{tabular}
\end{table}

\subsection{Ablation Analysis}
\label{sec:ablation}

Table~\ref{tab:ablation} evaluates which score components are responsible for the improvement of \method{}. Results are averaged over the eight dataset pairs in Table~\ref{tab:main}, with each dataset averaged over five seeds. Starting from the reproduced baseline score, unnormalized and Euclidean prototype distances give limited or inconsistent gains. Using the diagonal Mahalanobis prototype improves the mean \auc{} by $+1.27$ points and improves all eight datasets. This shows that normal-mode deviation is the main added signal.

Reliability gating further improves the prototype correction. The row with $q$ is the linear gate special case of \method{}, equivalently setting $\gamma=1$, and it raises the average gain to $+1.91$ points. Allowing the selected curvature $q^\gamma$ gives the final score and reaches $+2.03$ points. Shuffling the reliability values weakens the result, while inverse reliability reduces the mean below the baseline. These controls show that the direction of the gate matters. Keypoint confidence acts as reliability control for prototype evidence, not as an arbitrary scaling factor.

\begin{table}[t]
\centering
\caption{Ablation analysis of \method{} with mean \auc{} (\%) and Gain. Each row is averaged over the eight dataset pairs and five seeds. Positive datasets count how many dataset-level means improve over the corresponding baseline score.}
\label{tab:ablation}
\footnotesize
\begin{adjustbox}{max width=\linewidth}
\begin{tabular}{lccc}
\toprule
Variant & Mean \auc{} (\%) & Gain & Positive datasets \\
\midrule
Baseline & 79.68 & -- & -- \\
Baseline + unnormalized prototype distance & 79.55 & $-0.14$ & 2/8 \\
Baseline + Euclidean prototype & 79.84 & $+0.15$ & 5/8 \\
Baseline + diagonal Mahalanobis prototype & 80.95 & $+1.27$ & 8/8 \\
Baseline + prototype $\times q$ (linear gate, $\gamma=1$) & 81.60 & $+1.91$ & 8/8 \\
Baseline + prototype $\times$ shuffled reliability & 80.92 & $+1.24$ & 8/8 \\
Baseline + prototype $\times$ inverse reliability & 79.08 & $-0.60$ & 2/8 \\
\method{} = Baseline + prototype $\times q^\gamma$ & \textbf{81.72} & $\mathbf{+2.03}$ & 8/8 \\
\bottomrule
\end{tabular}
\end{adjustbox}
\end{table}

\subsection{Score Rank Distributions}
\label{sec:score-analysis}

\begin{figure}[!t]
\centering
\includegraphics[width=0.98\linewidth]{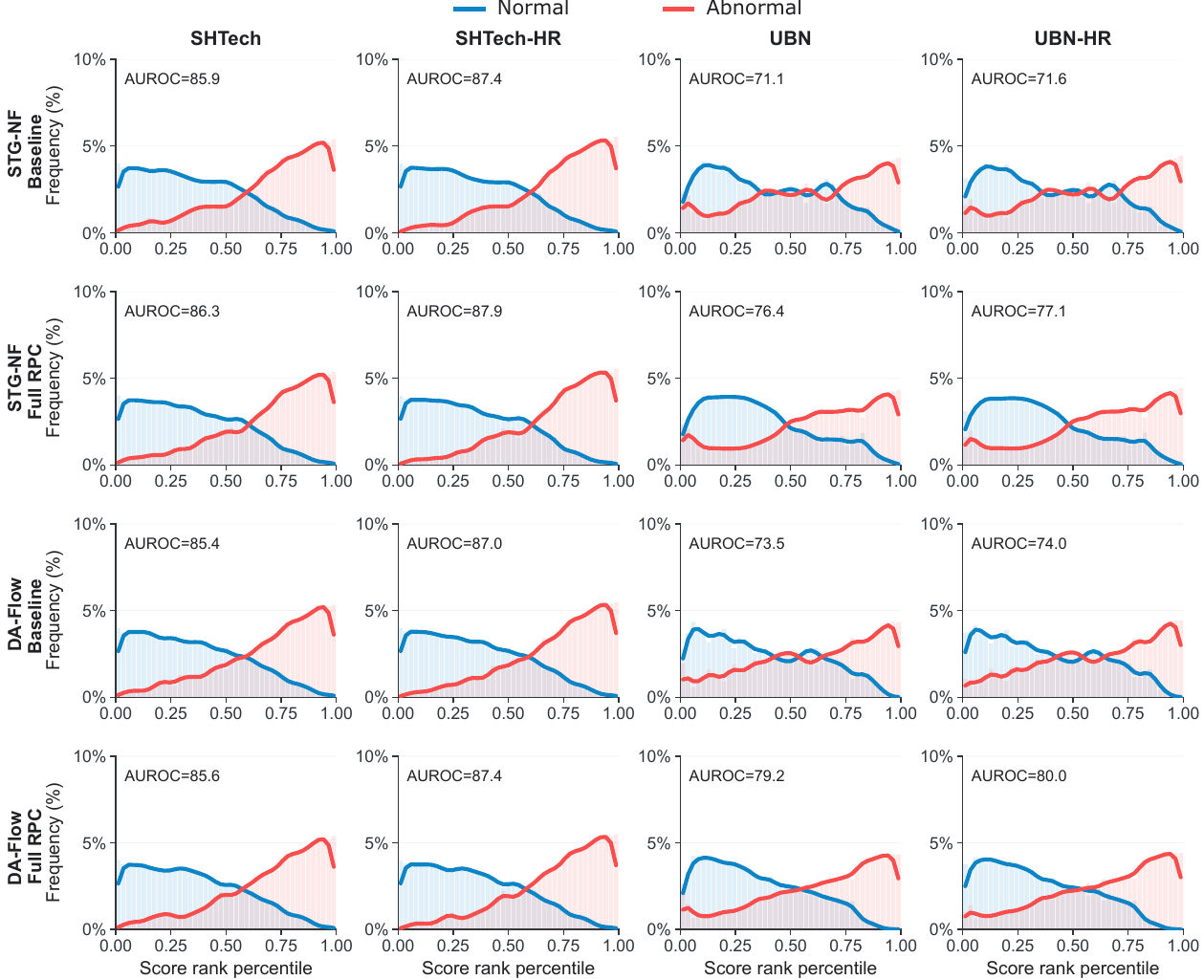}
\caption{Rank distributions of frame scores before and after calibration. Scores are converted to rank percentiles within each method for visualization. A better ranking places normal frames toward the left and abnormal frames toward the right. Columns show SHTech, SHTech-HR, UBN, and UBN-HR. Rows show \stgnf{} Baseline, \stgnf{}+\method{}, \daflow{} Baseline, and \daflow{}+\method{}. Bars are proportional histograms, and solid curves are Gaussian-smoothed profiles. All \auc{} values are computed from the original frame scores \((seed=42)\).}
\label{fig:score-distribution}
\end{figure}

Figure~\ref{fig:score-distribution} compares the rank distributions of frame scores before and after calibration. For each method and dataset, scores are converted to rank percentiles only for visualization. A better ranking places normal frames toward low percentiles and abnormal frames toward high percentiles. After applying \method{}, the abnormal distribution moves further to the high percentile region, while the normal distribution remains concentrated at lower percentiles. This pattern is clearest on UBN, where both frozen backbones also show larger gains in Table~\ref{tab:main}. On SHTech and SHTech-HR, the visual shift is smaller because the baseline scores already rank many abnormal frames near the high percentile end. This visual trend is consistent with the larger UBN gains in Table~\ref{tab:main}, and with the smaller gains on SHTech where the baseline rank distribution is already more separated.

\subsection{Analysis by Pose Reliability}
\label{sec:reliability-strata}

Table~\ref{tab:reliability} evaluates whether the reliability gate behaves consistently across different levels of pose quality. For each dataset pair, test frames are sorted by segment-level reliability and divided into low, middle, and high reliability groups. We then compute frame-level \auc{} within each group for the baseline score, the prototype score, and the full \method{} score.

The full score improves over the baseline score in all three groups, with gains of $+3.88$, $+3.43$, and $+2.51$ points from low to high reliability. It also improves over the prototype score in all groups. The additional gain over prototype calibration is largest in the low reliability group, where the full score adds $+2.36$ points. The margin becomes smaller in the middle and high reliability groups, where prototype evidence is already more stable. This pattern supports the intended role of reliability gating. It gives the strongest correction when pose observations are less reliable, while preserving the benefit of prototype evidence when pose reliability is higher.

\begin{table}[t]
\centering
\caption{\auc{} (\%) by pose reliability group, averaged over the eight backbone--dataset pairs. Test frames are divided into low, middle, and high reliability groups by segment reliability \((seed=42)\).}
\label{tab:reliability}
\footnotesize
\begin{tabular}{lccccc}
\toprule
Reliability split & Baseline & Prototype & Full \method{} & Full $-$ Baseline & Full $-$ Prototype \\
\midrule
Low & 75.18 & 76.70 & 79.06 & $+3.88$ & $+2.36$ \\
Mid & 79.99 & 82.81 & 83.42 & $+3.43$ & $+0.61$ \\
High & 76.92 & 79.15 & 79.43 & $+2.51$ & $+0.28$ \\
\bottomrule
\end{tabular}
\end{table}

\subsection{Robustness to Controlled Pose Perturbations}
\label{sec:robustness}

To assess robustness to test-time pose degradation, we perturb only the test skeletons and evaluate the resulting frame-level ranking. The perturbations cover four common pose failures, including confidence corruption, coordinate jitter, keypoint dropout, and tracking switches. Each type is applied at low, middle, and high severity for every backbone and dataset. Table~\ref{tab:noise_protocol} shows the perturbations.

\begin{table}[t]
\centering
\caption{Controlled pose perturbations. Coordinate jitter uses Gaussian noise in normalized coordinate units. Tracking switch replaces a contiguous track segment with another skeleton from the same clip.}
\label{tab:noise_protocol}
\footnotesize
\begin{tabular}{lccc}
\toprule
Noise type & Low & Mid & High \\
\midrule
Confidence corruption & Scale by 0.8 & Scale by 0.5 & Randomize confidence \\
Coordinate jitter & $\sigma=0.01$ & $\sigma=0.03$ & $\sigma=0.05$ \\
Keypoint dropout & Drop $10\%$ & Drop $30\%$ & Drop $50\%$ \\
Tracking switch & $20\%$ frames & $40\%$ frames & $60\%$ frames \\
\bottomrule
\end{tabular}
\end{table}

\begin{figure}[t]
\centering
\includegraphics[width=\linewidth]{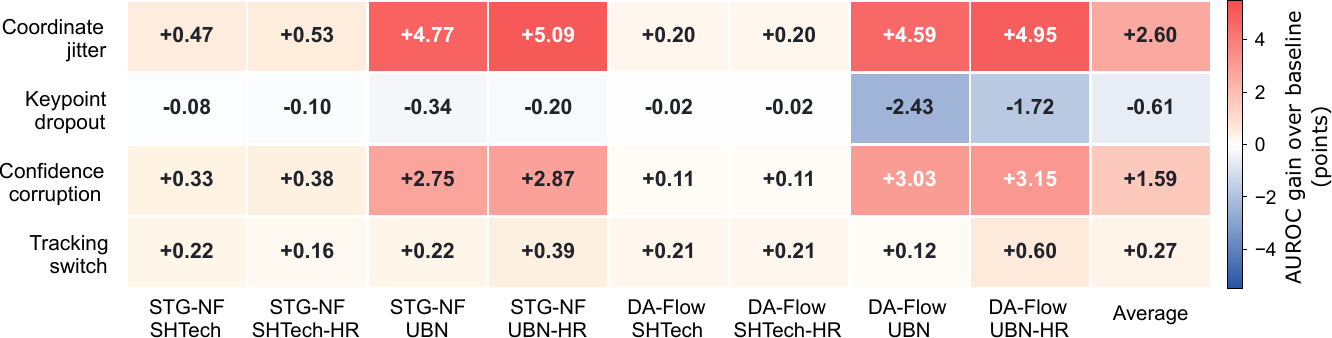}
\caption{Noise robustness gain of the full \method{} score over the baseline score. Each cell reports the \auc{} gain in points, averaged over low, middle, and high severity for the corresponding perturbation type. The rightmost column reports the mean across all eight settings. Red cells indicate improved robustness over the baseline score, while blue cells indicate lower performance.}
\label{fig:noise_delta}
\end{figure}

Figure~\ref{fig:noise_delta} reports the \auc{} difference between the full \method{} score and the baseline score, averaged over low, middle, and high severities for each perturbation type. The calibrated score gives the clearest gains under coordinate jitter and confidence corruption, with average improvements of $+2.60$ and $+1.59$ points. Tracking switches are close to neutral on average, with a small positive difference of $+0.27$ points. Keypoint dropout is the main failure case, with an average difference of $-0.61$ points. Unlike jitter or confidence corruption, dropout removes joints from the pose structure itself. This can distort the latent geometry used by both the flow model and the prototype distance. The reliability gate can reduce the authority of the prototype term, but it cannot recover the missing body structure. As a result, the added prototype evidence may lose discriminative contrast under severe missing joints.

\subsection{Sensitivity and Runtime}
\label{sec:sensitivity-runtime}

\begin{figure}[t]
\centering
\includegraphics[width=\linewidth]{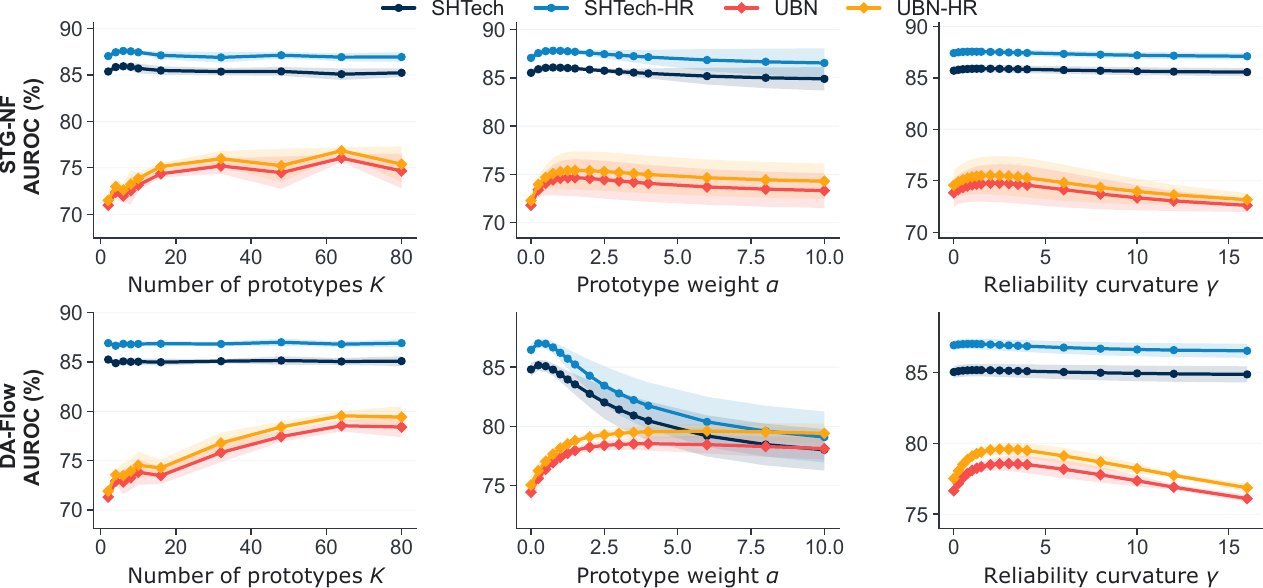}
\caption{Sensitivity of \method{} over five seeds for the number of prototypes $K$, prototype weight $\alpha$, and reliability curvature $\gamma$. Each curve varies one parameter while keeping the others fixed to the reported configuration. The vertical axis reports frame-level \auc{} (\%) and the shaded bands denote one standard deviation over five seeds.}
\label{fig:sensitivity}
\end{figure}

This experiment evaluates how \method{} behaves when one numerical parameter is varied at a time. We change the number of prototypes $K$, prototype weight $\alpha$, and reliability curvature $\gamma$, while keeping the other parameters fixed. Figure~\ref{fig:sensitivity} reports the mean \auc{} over five seeds with standard deviation bands.

The curves clarify what each parameter controls. Increasing $K$ first improves the coverage of normal motion patterns, especially on UBN, where normal behavior is more diverse. Once enough prototypes are available, further increasing $K$ brings limited gains. The $\alpha$ curves show how much prototype evidence should modify the baseline ranking, while the $\gamma$ curves show how strongly unreliable poses should suppress that evidence. These trends indicate that the selected parameters are not isolated optima, but fall in stable ranges that match the intended roles of the three components.

For runtime evaluation, \method{} adds no trainable parameters. The prototype bank is built once from normal training windows as an offline calibration step. At test time, the added computation is nearest-prototype scoring, reliability gating, and score fusion. The measured overhead varies across settings because the prototype count and latent dimensionality differ across configurations. The average overhead is 0.118 ms per frame, supporting \method{} as a lightweight post-hoc calibration module for frozen pose-flow detectors.

\subsection{Qualitative Case Studies}
\label{sec:case-study}

\begin{figure}[t]
\centering
\includegraphics[width=0.9\linewidth]{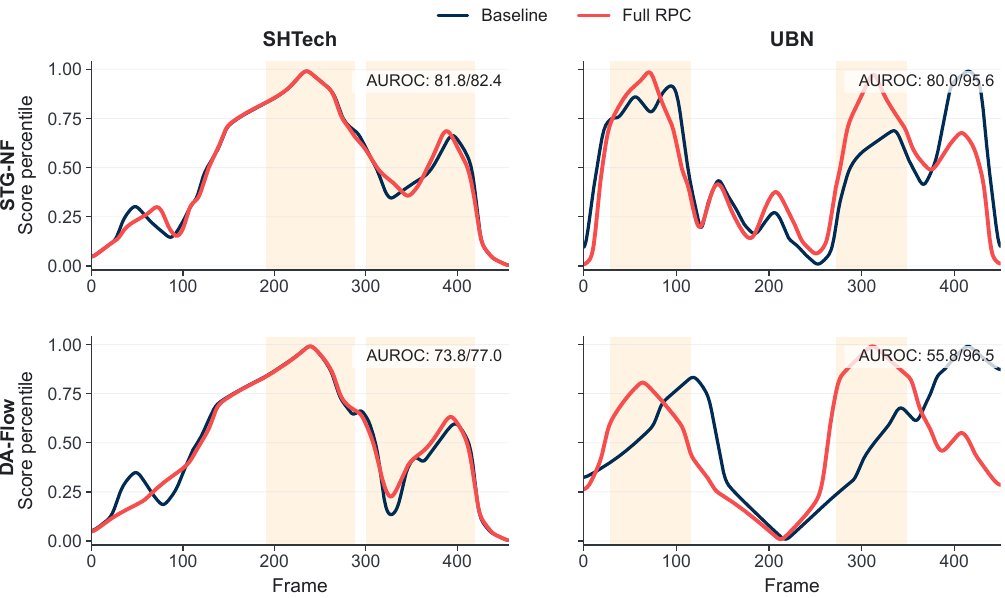}
\caption{Case studies on selected SHTech and UBN clips. Columns and rows are datasets and frozen backbones, respectively. Shaded regions mark abnormal intervals. Curves show frame-level anomaly score percentiles for baseline and \method{}. \auc{} pairs are ordered as baseline and \method{} and are computed from the original scores \((seed=42)\).}
\label{fig:case-detection}
\end{figure}

The case studies examine how the aggregate gains appear in individual clips. Figure~\ref{fig:case-detection} shows selected SHTech and UBN clips under the two frozen backbones. The figure compares the frozen baseline score with the final \method{} score.

On SHTech, the baseline score is already well aligned with the abnormal intervals, so \method{} mainly sharpens local responses near the shaded regions. On UBN, the baseline score is less consistent. The calibrated score raises responses inside abnormal intervals and suppresses several high responses in normal regions, producing clearer temporal separation. This two-sided change is important for frame-level ranking, because abnormal frames should move above normal frames rather than all scores increasing together. This indicates that \method{} improves the relative ordering of normal and abnormal frames, rather than simply increasing all scores in the clip.

\section{Discussion}
\label{sec:discussion}

\subsection{Interpreting Prototype Evidence}
\label{sec:discussion-prototype}

The role of prototype deviation is best understood as a ranking correction rather than as a new detector. The frozen likelihood score remains useful because it preserves the density evidence learned by the pose-flow model. The prototype term adds a different view. It asks whether the latent representation is close to empirical normal modes observed in training. This explains why replacing the flow score with standalone latent-space detectors is weaker than calibrating the original score. In heterogeneous scenes, especially UBN and UBN-HR, the nearest-prototype geometry provides additional separation because normal behavior is less well represented by a single scalar likelihood ranking.

\subsection{Reliability Gating as Evidence Control}
\label{sec:discussion-reliability}

The reliability analysis clarifies why the gate is useful. Its main effect is not to create a new confidence-based anomaly signal, but to decide when prototype geometry should be trusted. When reliability is low, prototype distance is more likely to mix behavioral deviation with pose estimation error. Gating therefore prevents this noisy geometric evidence from overwhelming the likelihood ranking. When reliability is high, the gate stays close to ungated prototype fusion, allowing the method to retain the benefit of normal-mode correction. This behavior supports the intended interpretation of reliability as evidence control.

\subsection{Failure Modes}
\label{sec:discussion-failure}
The controlled perturbation study reveals a clear boundary of score-level calibration. \method{} remains helpful when the pose is noisy but still structurally informative, such as under coordinate jitter or confidence corruption. In contrast, keypoint dropout removes body-structure information that both the likelihood score and the prototype distance depend on. Reliability gating can reduce the influence of questionable prototype evidence, but it cannot reconstruct missing geometry or repair the cached skeleton sequence.

This limitation is inherent to the frozen post-hoc setting. Once the cached pose track loses critical joints or body configuration, the latent representation may no longer contain the normal-motion geometry needed by the prototype bank. Improving this failure case would require better pose recovery, uncertainty-aware tracking, or additional cached cues beyond skeleton coordinates.

\subsection{Limitations}
\label{sec:limitations}

The first limitation is proxy fidelity. For SHTech and SHTech-HR, hyperparameters are selected without using test videos, trajectories, scores, or labels, but the proxy anomalies remain synthetic approximations of abnormal behavior. They cover common pose and trajectory degradations, not the full space of real abnormal events. For datasets without official validation splits, transparent proxy construction and candidate-space disclosure are therefore important for reproducibility.

The second limitation is information availability. \method{} operates only on cached skeleton tracks, frozen latent features, likelihood scores, and keypoint confidence. It does not modify pose extraction, tracking, representation learning, or window-to-frame aggregation. Consequently, it cannot recover missing body structure or use appearance and object cues that were never stored in the cache.

\section{Conclusion}
\label{sec:conclusion}

We presented \method{}, a post-hoc score calibration method for frozen pose-flow video anomaly detectors. By adding a reliability-gated nearest-prototype deviation to the standardized likelihood score, \method{} corrects the frozen ranking with empirical normal-mode structure while keeping the backbone, cached skeleton tracks, and evaluation pipeline unchanged. Experiments on two frozen pose-flow backbones and four datasets show consistent frame-level \auc{} improvements, and the analyses indicate that prototype deviation provides the main correction while reliability gating controls pose-dependent geometric evidence. The approach is lightweight and practical for cached pose-flow systems, but it remains bounded by the information present in the skeleton cache. Future work may extend the same post-hoc view to cached appearance, object, or tracking-uncertainty cues.

\bibliographystyle{unsrtnat}
\bibliography{references}

\end{document}